# Improving Financial Cybersecurity Through Fraud Transaction with Deep Learning-Based Advanced Models

Harish Kant Pathak
*Independent Researcher*
harish.pathak@gmail.com
*ORCID-0009-0004-4390-4170*

Rahul Kumar
*Independent Researcher*
logtorahul@gmail.com
*ORCID-0009-0004-4390-4170*

***Abstract*—The quick digitalization of financial services has greatly exposed financial operations to cyber threats. Conventional rule-based cybersecurity controls do not have the strength to protect against the changing attack schemes, such as fraud, identity theft, adversarial manipulation, and advanced persistent threats. The research presents a deep learning-based approach to detect fraudulent transactions by using the Fraud Detection Transactions Dataset. The proposed solution includes a large amount of data preparation, such as outlier detection, feature engineering, label encoding, feature scaling, and class imbalance control with SMOTE. There are two types of categorization models: artificial neural networks (ANNs) and wide-and-deep neural networks. Experimental results show that the Wide & Deep model outperforms the ANN in acc (0.9846), prec (0.9634), rec (0.9895), F1-score (0.9763), and ROC-AUC (0.99). Fraud patterns of complex nature can be detected by the linear and DL elements. In addition, an ablation study is applied to show the effects of the various optimizers, and their comparison with the current models proves the power of the developed approach. The results demonstrate that cybersecurity systems for financial transactions may greatly benefit from using the Wide and Deep architecture to enhance the precision and resilience of fraud detection.**

## I. INTRODUCTION

The financial ecosystem of the world has shifted to digital platforms, such as online banking, mobile payment systems, e-wallets, and cryptocurrencies. Although digital transformation brings benefits such as greater efficiency and accessibility to operations, it exposes a company to more cyber threats [1]. There is a wide variety of illegal behaviors that go under the umbrella phrase "banking fraud," including phishing, account takeover, credit card fraud, check fraud, and identity theft. Financial cyber fraud has become a major menace in the financial sector in the digital age, affecting individuals, companies, and governments around the world. Cybercriminals utilize sophisticated techniques to take advantage of flaws in financial systems, which may lead to significant financial losses, harm to one's reputation, and even legal consequences [2]. A growth in online transactions and the digitalization of financial services has brought attention to the need for reliable fraud detection systems.

Financial fraud involves a variety of strategies that defraud customers of their money, such as credit card fraud, money laundering, and insurance fraud [3]. Despite the significant breakthroughs that artificial intelligence (AI)-based fraud detection has shown, it has certain inherent limits [4]. Model performance may be adversely affected by a lack of generalizability, bias resulting from unbalanced datasets, and reliance on data quality. Adversarial attacks also make it more difficult to detect financial fraud as they imply that criminals manipulate the transaction data to avoid detection systems [5][6]. Overestimated false positive rates and long reaction times are consequences of conventional fraud detection systems' failure to adapt to evolving fraud tactics. These systems depend on rule-based algorithms and human investigations [7]. One possible solution could lie in evolving methods of detection, notably with the help of DL(Deep Learning) and ML (Machine Learning) [8]. These techniques can enhance the effectiveness of fraud protection systems since they can process large amounts of data at any given moment, detect intricate patterns, and evolve to new approaches of fraud [9]. DL is a branch of machine learning, and it is able to identify complex fraud patterns by applying multi-layered neural networks to derive abstract models using data. Models based on Convolutional Neural Networks (CNNs), Long Short-Term Memory (LSTM) networks, and Natural Language Processing (NLP) have shown promise for real-time fraud detection. This hybrid structure fuses a wide (linear) aspect of feature interaction memorization and a deep (nonlinear) aspect of generalization. Incorporating these factors can go a long way in enhancing the ability of the model to detect both simple and sophisticated trends in the data of financial transactions. The study's most significant contributions include the following:

- Developed a universal fraud detection model that included Synthetic Minority, feature engineering, and preprocessing. Using the SMOTE Over-sampling Technique to Fix Class Inequality.
- Designed and trained Artificial Neural Network (ANN) and Models for fraud classification using Wide and Deep Neural Networks.
- To determine how various optimizers impacted the model's performance, an optimizer ablation study was conducted.
- The model's performance was assessed using F1-score, ROC-AUC, recall, accuracy, precision, and other measures.

### *A. Motivation of the study*

A novel level of risk has been introduced into the financial sector by the rapid expansion of digital payment systems such as internet banking, mobile payments, and shopping online, and existing traditional rule-based methods of detecting fraudsters are not sufficient because they are too slow to adjust to the constantly changing fraud behavior and too powerless to work with heavy data volumes and imbalanced data. Thus, intelligent and scalable solutions that can correctly identify complex and dynamic fraud behaviors in real-time are highly needed, hence the desire to apply advanced deep learning methods in this study.

## B. *Organization of the Paper*

An overview of the project is as follows: In Section II provides a synopsis of the relevant literature and current methods for fraud in financial transactions. The suggested approach is outlined in Section III. The findings, analysis, comparison with existing models, and recommendations for future study are all included in Section IV. Section V concludes the investigation.

# II. LITERATURE REVIEW

Recent research has examined the effectiveness of ML and DL approaches in identifying instances of financial fraud. S. Mallikarjunaiah et al. (2025) suggested method would combine CNN and LSTM networks to find intricate fraud patterns in transaction data, while SVM will be effective when applied to structured data. The system has an accuracy of 96, which is very high compared to the traditional techniques like random forest and SVM, thus it is very strong in detecting real-time fraud [10]. Ebnezer et al. (2025) proposed DL method utilizes Bi-LSTMs to identify fraudulent financial transactions in real-time. The model recorded high performance with an acc of 91.8, F1score 90.5 and PR-AUC of 0.94 compared to the traditional ML algorithms like the LR, RF, and XGBoost [11]. Additionally, K. P. Kumar et al. (2025) proposed a DL model that employs an Attention Mechanism and Bi-LSTM to increase precision of identifying fraud in Ethereum transactions. On a Kaggle Ethereum dataset, the model achieved 97% acc, 97% prec, 97% rec, and a 97% F1score, much higher than existing works [12].

R. Sharma and S. Gupta (2024) explored the potential of improving fraud detection in online banking with the help of DL algorithms. The LSTM model outperformed others in terms of prec, acc, F1score, recognition, and AUC-ROC of 0.936, 95.8 and 82.7, respectively. The CNN model demonstrated the following results: 95.2% accuracy, 84.1% precision, and 77.8% recall, in terms of accuracy, precision, recall, F1score, and AUC, respectively [13]. In the study conducted by R. Sharma and A. Sharma (2024), provided significant volume of transactional data in their research, and a combination of supervised and unsupervised learning approaches was employed to determine problematic ones. This method also automates data cleansing. As shown by its 0.972 AUC, 93.7% sensitivity, 97.5% specificity, and 95.8% accuracy, the RNN architecture outperforms the CNN model [14]. N. R. Shanbhog et al. (2024) compared advanced tools of analysis in the effective separation of legitimate and deceptive patterns in financial data. The three models - XGBoost, ANN, and R-GCN demonstrated good results, with accuracies 97.0%, 94.0%, 98.5% respectively [15].

P. Y. Prasad et al.'s (2023) main objective is to develop a ML-based fraud detection solution. The basis of algorithmic frameworks like XGBoost, RF, SVM, DT, and LR is machine learning. The algorithms' output is impacted by the F1score, rec, acc, and prec. Values of 99.9%, 85.71%, 93%, and 98% have been used to modify the Area Under Curves (AUC) [16]. Current fraud detection techniques based on CNNs, LSTMs, and other sophisticated models tend to target more sophisticated patterns however they do not merge the linear and non-linear relations well. The problem of class imbalance is not well addressed in much research, and as a consequence, the findings are skewed since they are not examined using a variety of criteria. In addition, there has been a lack of research on optimizer analysis and model comparison, which highlights the need for a stronger and more versatile model to improve fraud detection overall.

# III. METHODOLOGY

Methodology and process for the suggested fraud detection system are shown in Fig. 1. Preprocessing the data includes tasks such as identifying outliers, engineering features, encoding labels, and scaling features. The procedure begins with a dataset of fraud detection transactions. To correct the class imbalance, the data is processed and then split into a training set and a test set using SMOTE. Then ANN and Wide and Deep models are applied to classification. The models tested according to recall, precision, accuracy, F1score, and ROC-AUC, and ablation study conducted, which result in the final results.

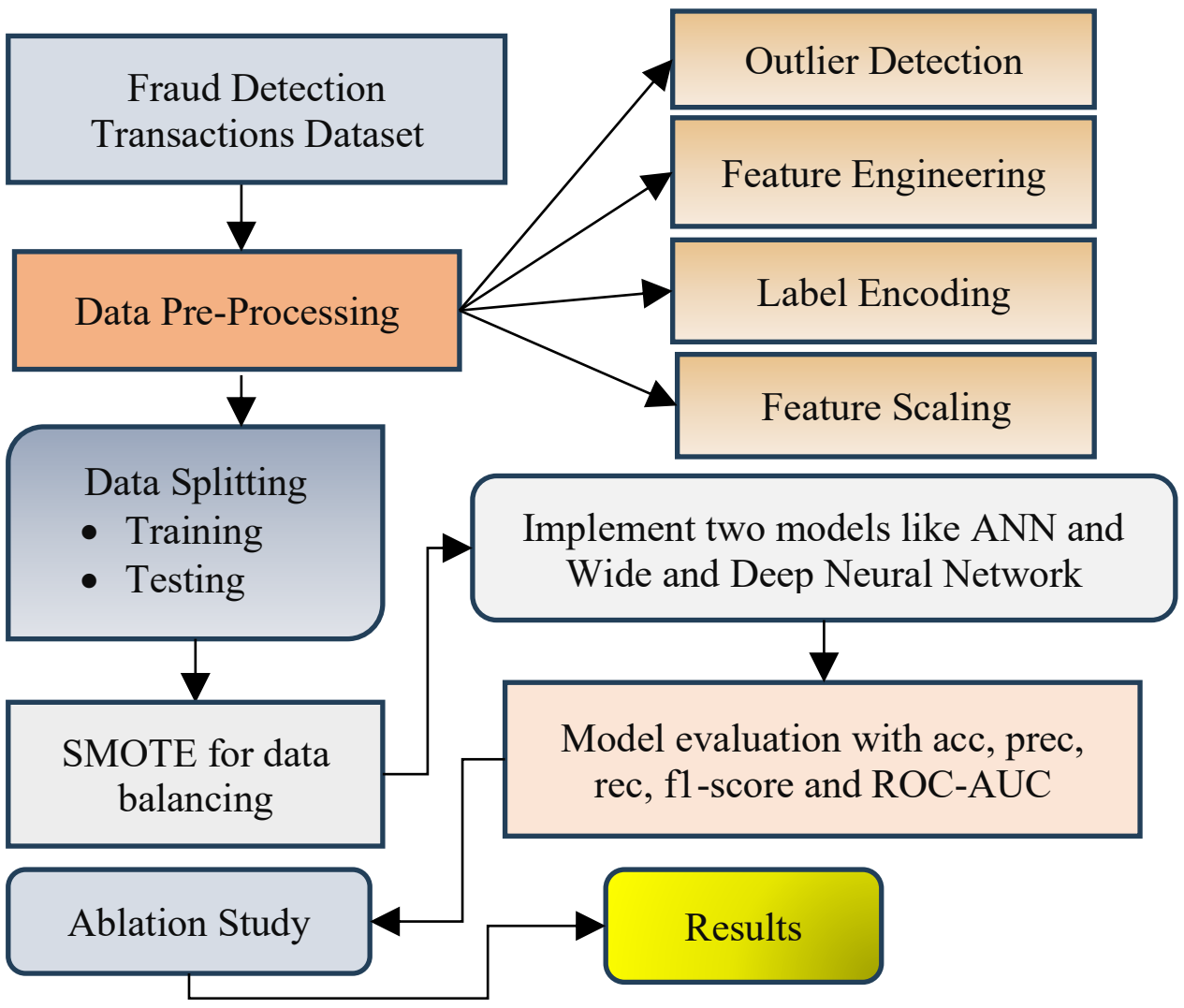

Fig. 1. Proposed Methodology for fraud detection system.

The proposed methodology steps and implementation strategy are elaborate below:

## A. *Data Collection and Visualization*

The dataset that utilized in this research is Fraud Detection Transactions Dataset available on Kaggle and includes 50,000 transaction records that have 21 features. The dataset shows a substantial disparity between genuine and fraudulent transactions, with 76% of the transactions being legitimate and 24% being fraudulent. There are numerical and nominal elements that indicate the user's behavior and transaction pattern, such as the amount of the transaction, the kind of the transaction, the account balance, the location, the merchant category, and the risk score. There is some visualization like heatmap, distribution graph etc. discuss below:

Fig. 2 shows the distribution of various significant variables in the fraud detection data, such as TransactionAmount, AccountBalance, IP AddressFlag, PreviousFraudulentActivity, DailyTransactionCount, AverageTransactionAmount (7 days), FailedTransaction count, CardAge, TransactionDistance, RiskScore, weekend indicator and fraud label. The plots display that a majority of the numerical characteristics including transaction amount and transaction distance are right skewed meaning that there are high-value outliers. Binary characteristics such as IPaddressFlag, PreviousFraudulentActivity, Is_Weekend and FraudLabel have a high rate of class imbalance with more normal cases than the fraudulent ones.

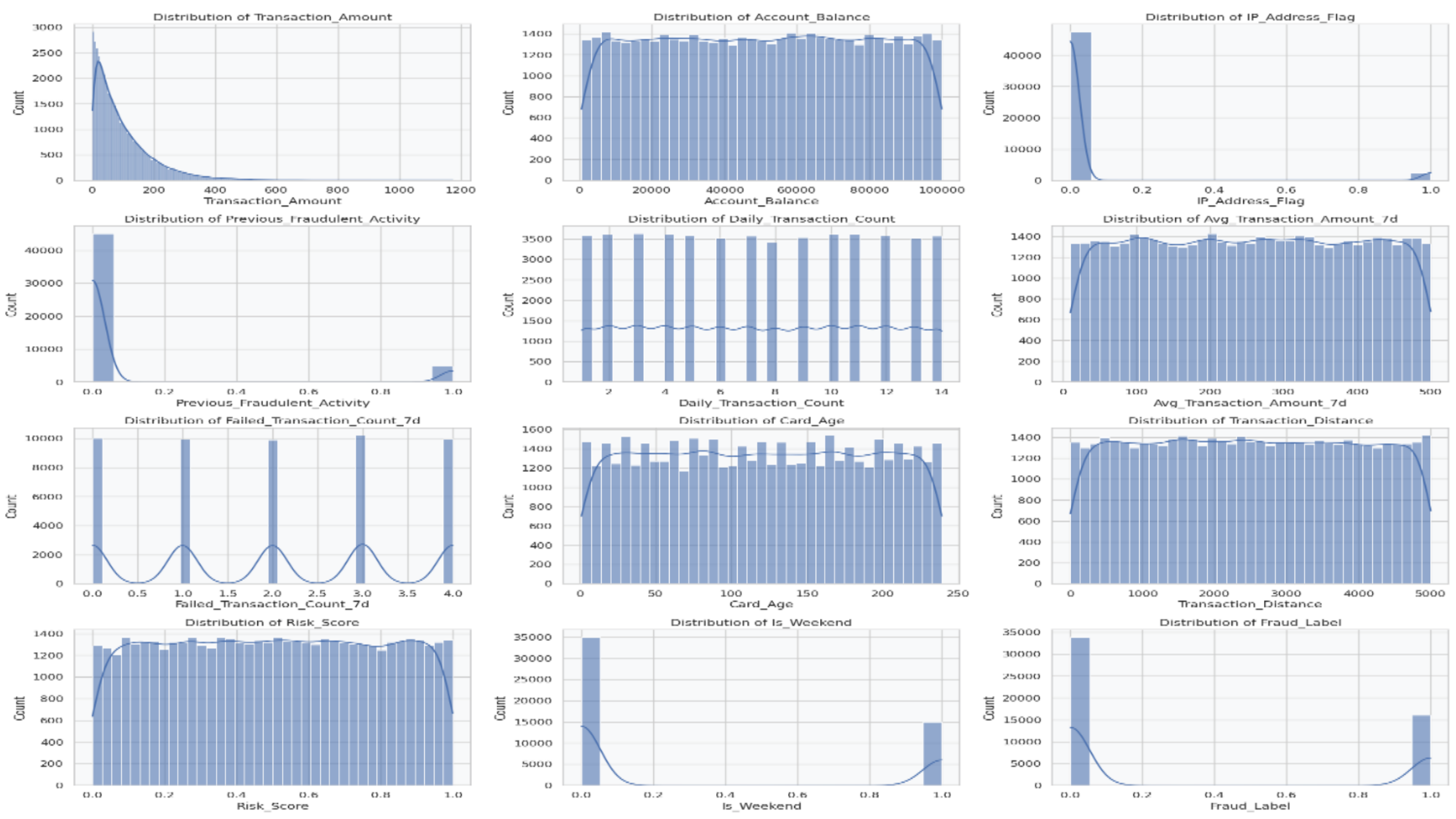

Fig. 2. Distribution of Key Transaction Features.

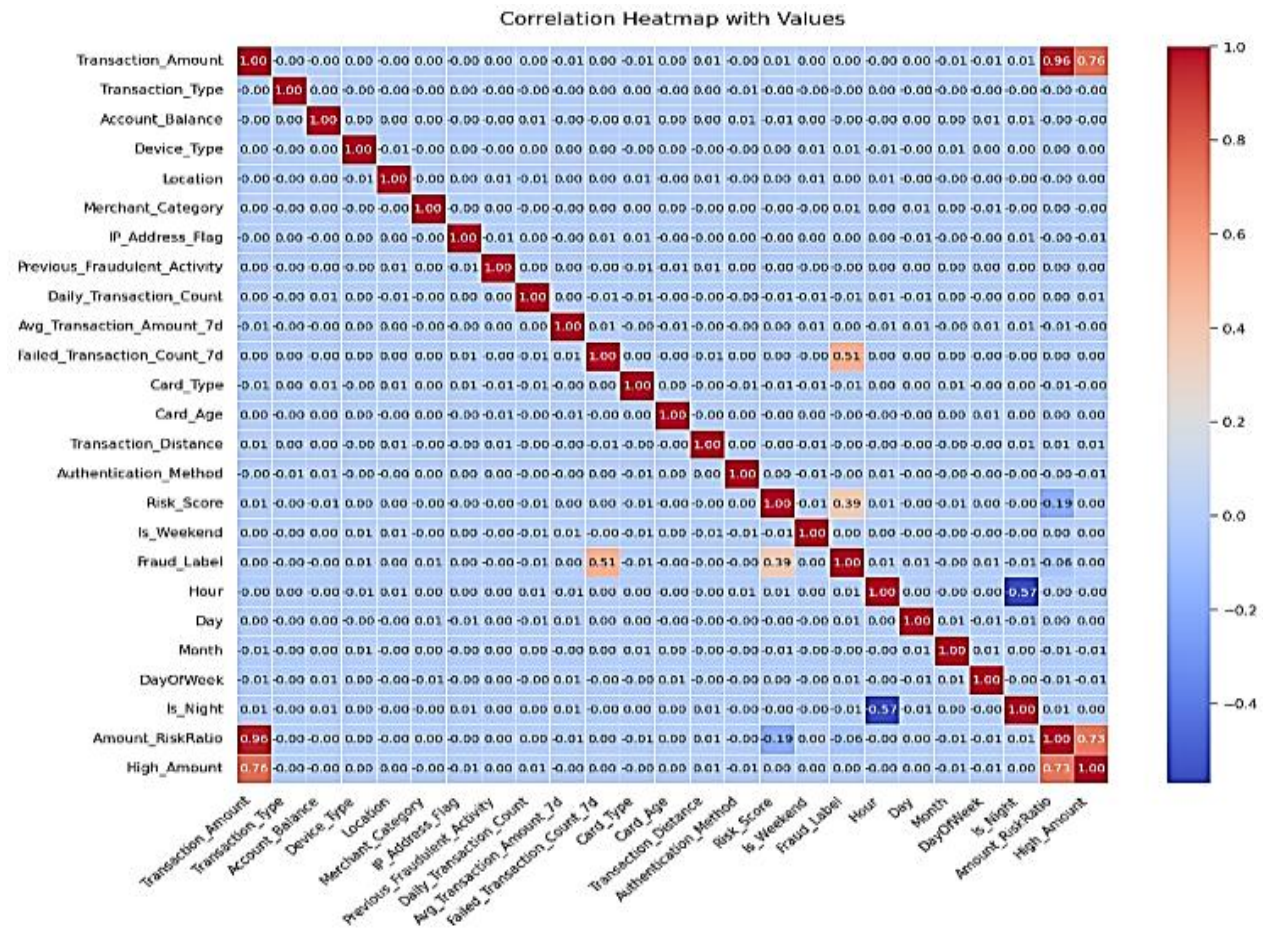

Fig. 3. Heatmap of Features and Values Correlation.

Fig. 3 heatmap gives correlation between all data features. Correlation is low between most features and this means that multicollinearity is low. There are increased correlations observed among some derived features such as Transaction Amount, Amount Riskratio and High amount. Most features are weakly correlated with the target variable Fraud_Label, which leads to the importance of complex models to ensure an ability to capture complex patterns.

## *B. Data Preprocessing*

Data is pre-processed to make sure it is prepared for model training. This comprised data cleaning, feature engineering and encoding of categorical variables as well as feature scaling. Also, outlier detection is done to detect extreme values of numerical features especially in the variables of Transaction_Amount and other associated variables.

## *C. Feature Engineering*

The feature engineering is done to add new meaningful features to the data based on the existing attributes. The amount of risk ratio (Transaction Amount)/(Risk Score + 1) is a continuous feature Amount Risk Ratio which is formed to provide the relative intensity of risk in a transaction without dividing by zero. Also, a binary attribute High_Amount is added, to reveal the high-value transactions, with the transactions beyond the 90th percentile of the Transaction_Amount called 1 and the rest 0. In addition to temporal features that have been previously extracted (i.e., Hour, Day, Month, DayOfWeek, and Is_Night), they are engineered features that make the model more effective in detecting latent patterns of fraud and lead to making the model better in terms of classification.

## *D. Label Encoding for Categorical Columns*

The Label Encoding is used to transform categorical features into numerical values. Under this technique, a different integer is given to each distinct category under a feature. The data incorporate features such as Transaction Type, Type of device, Location, Merchant Category, Card Type and method for authentication and verification and verify that data is structured properly for model training.

## *E. Feature Scaling*

Feature scaling is carried out using the StandardScaler approach to standardize the numerical data such that the model represents each variable equally. The model's overall performance and convergence are enhanced by standardization, which comprises changing the data to a mean of 0 and a StandardDeviation of 1. The transformation can be defined as Equation (1):

$$Z = \frac{X-\mu}{\sigma} \tag{1}$$

Where, $X$ is original FeatureValue, $\mu$ is mean of the feature and $\sigma$ is StandardDeviation of the feature.

## *F. Data Splitting*

The dataset is stratified-sampled and split 75:25 across the training and testing sets to keep the class distribution constant.

There are 37,500 samples with 24 features in the training set and 12,500 samples with 24 characteristics in the testing set.

### *G. Handling Class Imbalance*

The significant imbalance in the classes included in the dataset is addressed by the SMOTE. SMOTE is a method for balancing data that improves models' ability to handle imbalanced classification issues by creating synthetic minority examples via interpolation between existing instances.

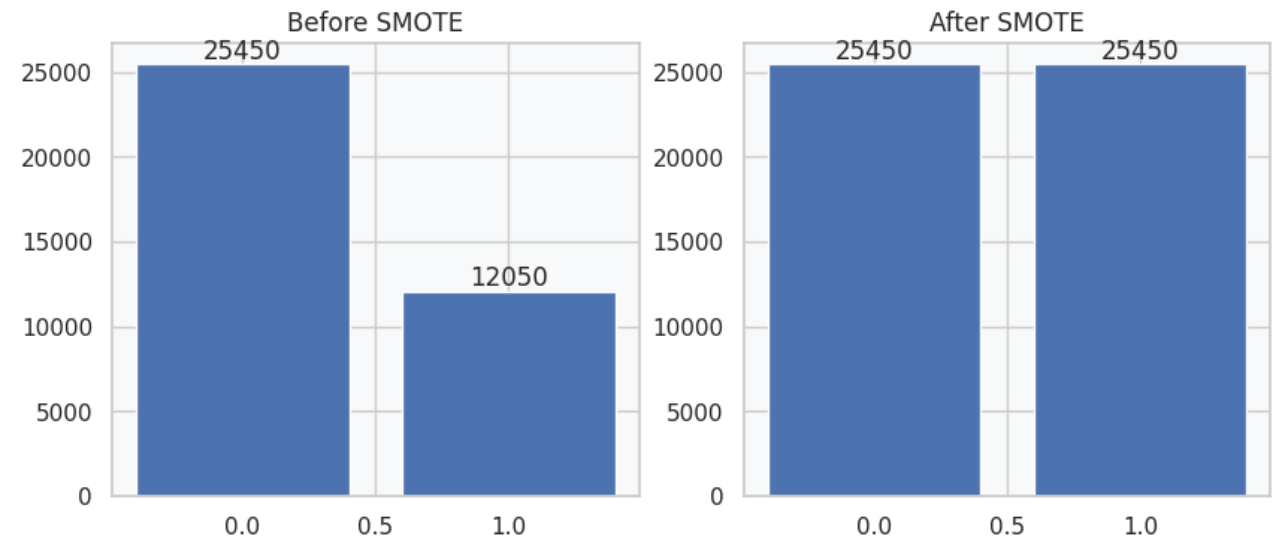

Fig. 4. Distribution of classes before and after SMOTE.

Fig. 4 the left plot, the initial imbalanced data is presented, in which the non-fraud group (0) has 25,450 samples and the fraud group (1) has 12,050 samples. The right plot shows the balanced dataset, which is balanced after the use of SMOTE in both classes (0 and 1) 25,450 samples of each of the classes are equally represented, which means that the model is better trained, and there is less bias with the majority class.

### *H. Proposed Models for fraud detection*

The DL models include the ones listed below.

#### *1) Artificial Neural Network (ANN)*

A sequential architecture with layers of fully connected architecture is created to generate an ANN to detect fraud [17]. After an input layer to identify complex patterns in the data, the model consists of three hidden layers with ReLU activation, each having 128, 64, and 32 neurons. There are dropout layers (0.1 and 0.2) that followed the first two hidden layers to avoid overfitting. Binary classification uses the sigmoid activation function on the output layer. A BatchSize of 32 and 30 epochs is applied with a validation split of 20% to train.

#### *2) Wide and Deep Neural Network*

The Wide and Deep Neural Network uses memory and generalizability by modeling a deep neural network and a long linear framework as a fraud detection model. The wide component uses linear activation and a dense layer of 64 neurons to learn basic feature interactions. In contrast, the deep component employs ReLU activation in multiple dense layers of 128, 128, and 64 neurons to learn complex patterns. To further stabilize and regularize the model, it also makes use of BatchNormalization and dropout (0.1). The two parts are joined through concatenation and binary classification through a sigmoid output layer is used. They used the AdamOptimizer to train their model, using an initial learning rate of 0.001, BinaryCross-entropy loss as their evaluation measure, a BatchSize of 64, 50 epochs, and a valid test size of 50%.

#### *3) Experimental setup and Model Evaluation*

The following components make up the computational ecosystem: Powerful hardware: Includes a 32GB RAM stick, an Intel Core i9 CPU, and a 16GB NVIDIA GPU. The required software packages are TensorFlow 2.0, Scikit-Learn 0.24, Pandas, NumPy, and at least Python 3.8%. A wide range of evaluation metrics is utilized to evaluate how well fraud detection systems perform. When comparing all anticipated frauds to the proportion of accurately expected frauds, it gets the forecast's accuracy; when comparing this percentage to all projected frauds, it gets the forecast's precision. The sensitivity or recall of a model is a measure of its ability to identify real instances of fraudulent transactions. Rec and prec are both taken into account by the F1score. To evaluate a model's capabilities in class distinction across different thresholds, the ROC-AUC is a helpful statistic. All of them are computed using Equations (2) through (5):

$$Acc = \frac{TP+TN}{TP+FP+TN+FN} \quad (2)$$

$$Prec = \frac{TP}{TP+FP} \quad (3)$$

$$Rec = \frac{TP}{TP+FN} \quad (4)$$

$$F1-Score = 2 * \frac{(Precision*Recall)}{Precision+Recall} \quad (5)$$

Here, TP indicates that fraud cases have been appropriately identified, TN indicates that genuine transactions have been accurately classified, FP indicates that fraud instances have been reported incorrectly, and FN indicates that fraud cases have not been detected.

## IV. Result Analysis and Discussion

Table I displays the results of comparing the suggested ANN and Wide & Deep Neural Network models' performance on the fraud detection dataset. The results demonstrate that when comparing the two models, the Wide & Deep Neural Network model outperforms the ANN in terms of the majority of assessment criteria, including acc (0.9846), prec (0.9634), rec (0.9895), and the F1score (0.9763). Both of them are good at classification having the same ROC-AUC at 0.9990. The ANN may have a little higher recall (0.9983), but the Wide and Deep model outperforms it in all three metrics, making it the superior fraud detector.

TABLE I. Performance of Proposed Models for Fraud Transactions using Fraud Detection Transactions Dataset

| Matrix | ANN | Wide & Deep Neural Network |
|---|---|---|
| Accuracy | 0.9667 | 0.9846 |
| Precision | 0.9074 | 0.9634 |
| Recall | 0.9983 | 0.9895 |
| F1 Score | 0.9507 | 0.9763 |
| ROC-AUC | 0.9990 | 0.9990 |

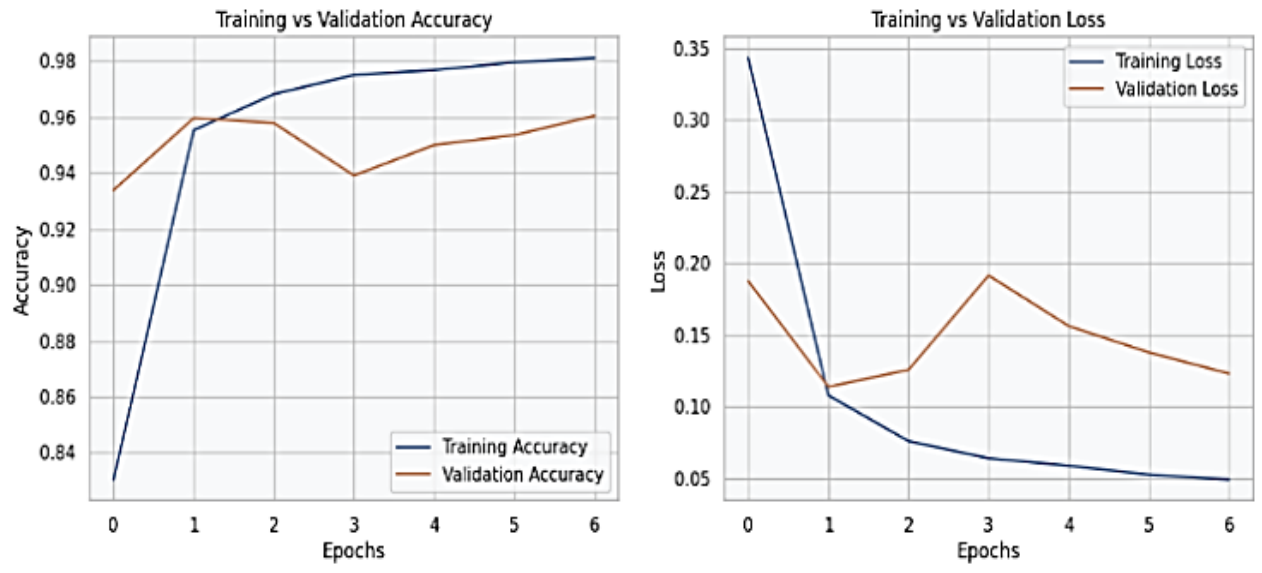

Fig. 5. Accuracy and loss graph for ANN model.

Fig. 5 displays the ANN model's accuracy and loss throughout epochs during training and validation. The accuracy plot shows that there is a gradual increase in training accuracy with constant changes in validation accuracy which is high and indicates good learning. The loss plot depicts steady reduction in training and slow reduction in validation loss following small fluctuations, which suggests successful training with little overfitting.

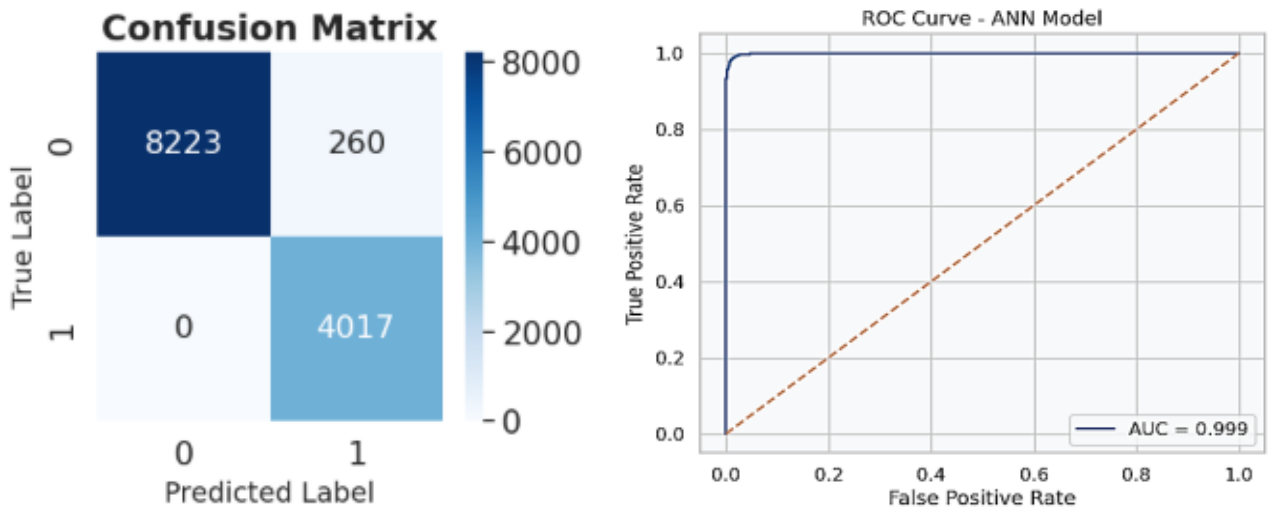

Fig. 6. ANN Model's Confidence Matrix and ROC Curv.

Fig. 6 displays the ANN model's ROC and confusion. The Confusion Matrix presents 8223 TN, 4017 TP, 0 FP, and 260 FN, which means that the classifier has a good performance with minimal mistakes. According to the ROC curve, the model has an impressive AUC of 0.999, which means it can tell the difference between good and bad transactions.

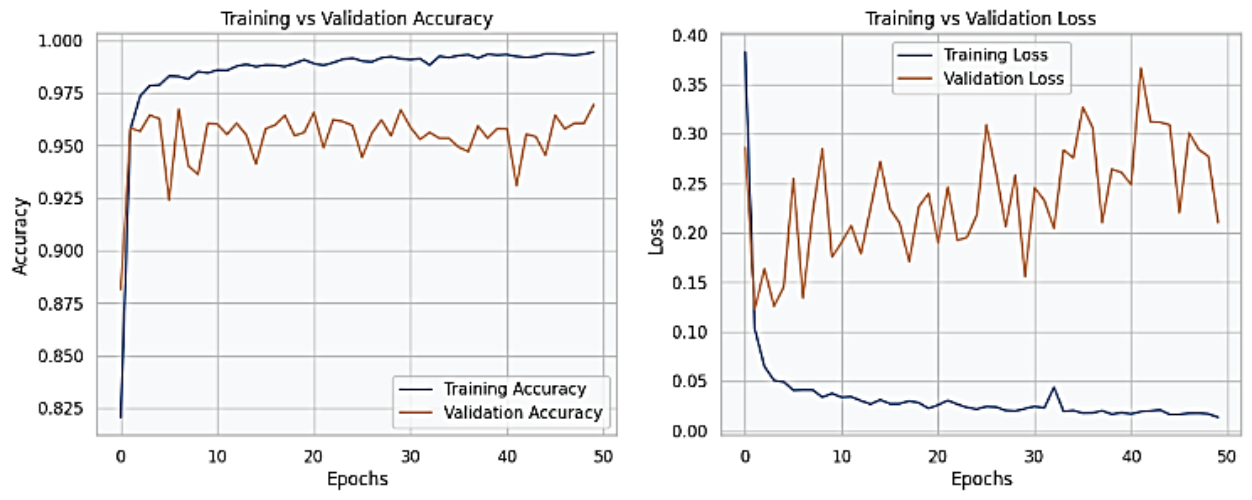

Fig. 7. Accuracy and loss graph for Wide & Deep Neural Network model.

Fig. 7 displays the Wide & Deep model's training and validation accuracies as well as its losses relative to epochs. The training accuracy rapidly attains high levels whereas the validation accuracy steadily attains low variations. The training loss is low, but validation loss is also not constant, which means that it is a high-learning model with minimal overfitting but with an overall good performance.

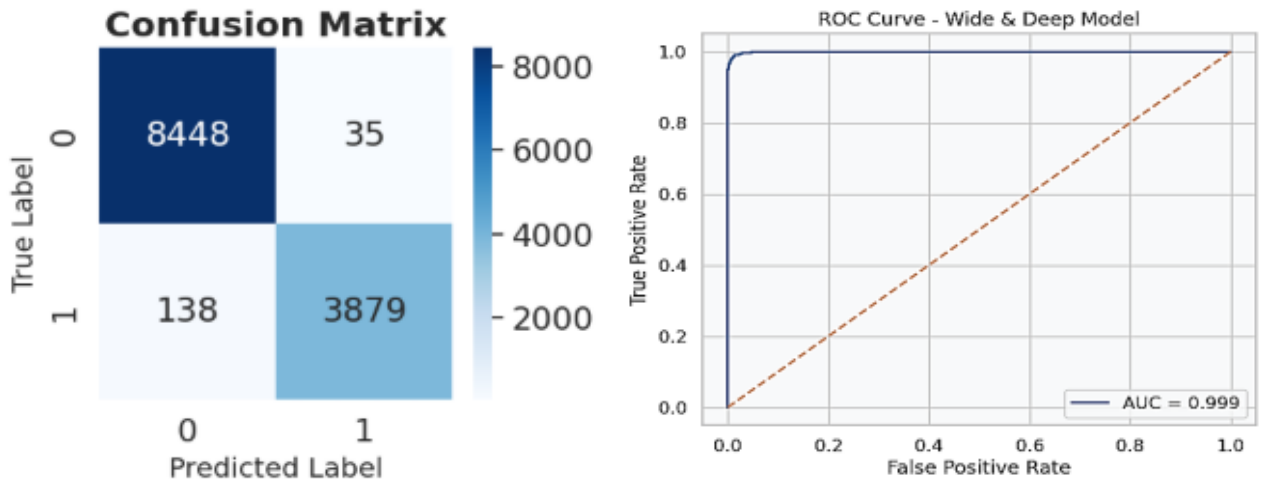

Fig. 8. Confusion matrix and ROC of Wide & Deep Neural Network model.

The Confusion Matrix and ROC curve for the Deep and Wide models are shown in Fig. 8. According to the Confusion Matrix, the low-error classification accuracy is represented by 8448 TN, 3879 TP, 138 FP, and 35 FN. The AUC of 0.999 for the ROC indicates that it is very good at differentiating between legitimate and fraudulent transactions.

## A. *Ablation Study*

Table II shows the results of Wide and Deep model involving the comparative performance of various optimizers using the optimizer ablation study. The findings show that SGD (lr = 0.01, momentum = 0.9) gives the most overall performance with highest acc (0.9649), prec (0.9895), F1score (0.9737), MCC (0.9220) and Kappa (0.9210). RMSprop is also a competitor to balanced precision and recall, with Adam also showing strong performance, Adam with slightly reduced performance. Adamax presents relatively weaker outcomes of all optimizers. SGD, in general, is a much better optimizer of this model, and it is more stable and generalizes better. Additionally, by using this Wide and Deep design, it may take advantage of stable learning dynamics, which allows for the identification of both linear and nonlinear correlations without overfitting the neural network model's training data..In practical terms, this makes the model more reliable and effective since the implementation of SGD provides greater robustness in its operation.

TABLE II. OPTIMIZER ABLATION STUDY FOR WIDE AND DEEP MODEL PERFORMANCE

| Optimizer | Acc | Prec | Rec | F1 | MCC | Kappa |
|---|---|---|---|---|---|---|
| Adam (lr = 0.001) | 0.9567 | 0.9738 | 0.9620 | 0.9679 | 0.9018 | 0.9017 |
| SGD (lr = 0.01, momentum = 0.9) | 0.9649 | 0.9895 | 0.9584 | 0.9737 | 0.9220 | 0.9210 |
| RMSprop (lr = 0.001) | 0.9641 | 0.9731 | 0.9740 | 0.9735 | 0.9179 | 0.9179 |
| Adamax (lr = 0.002) | 0.9592 | 0.9820 | 0.9573 | 0.9695 | 0.9085 | 0.9079 |

## B. *Comparative Study and Discussion*

Table III presents a comparison of the proposed and existing models' performance across various datasets. The results show that the suggested Wide and Deep Neural Network performs best overall in the dataset of fraud detection transactions, with greatest acc (0.9846), prec (0.9634), rec (0.9895), and F1score (0.9763), followed by the ANN model. The ANN model also performs well, achieving high recall (0.9983) and a balanced F1-score of 0.9507.

TABLE III. FRAUD TRANSACTION MODEL PERFORMANCE COMPARISON OF PROPOSED AND EXISTING MODELS

| Models | Acc | Prec | Rec | F1 |
|---|---|---|---|---|
| ANN | 0.9667 | 0.9074 | 0.9983 | 0.9507 |
| Wide & DNN | 0.9846 | 0.9634 | 0.9895 | 0.9763 |
| NB[18] | 0.9778 | 0.0419 | 0.8469 | 0.1163 |
| DT [19] | 0.95 | 0.909 | 0.919 | 0.928 |
| Logistics regression[20] | 0.92 | 0.04 | 0.76 | 0.08 |
| Random Forest[20] | 0.96 | 0.09 | 0.97 | 0.17 |
| XGBoost[21] | 0.926 | 0.958 | 0.902 | 0.929 |
| SVM[22] | 0.957 | 0.669 | 0.814 | 0.734 |

When it comes to metrics, the decision tree and XGBoost models have performed around average, but the Support Vector Machine model has done really well, with respectable recall and accuracy. Unfortunately, although having respectable recalls and accuracies, the random forest, logistic regression, and Naive Bayes models have all failed miserably when it comes to precision.

## C. *Areas for Improvement*

Although the suggested models have shown good results, there are some aspects that can be enhanced in future studies. The present research has been based on one data set, which might not be generalizable to other real-life financial systems. It is also possible to add real-time streaming data to the models to have live fraud detection. Also, to be useful in financial institutions, it is necessary to enhance the model interpretability. Ensemble or hybrid models would also increase performance and robustness. Lastly, looking into the complex methods of dealing with extreme imbalance classes and adversarial fraud patterns may enhance the stability of the system setup in changing conditions.

## V. CONCLUSION AND FUTURE WORK

This research provided a thorough framework for fraud detection that uses Wide & Deep Neural Network architecture. It aims to overcome the constraints of standard deep learning methods that only use one route. The implementation of the proposed solution showed that the framework is capable of detecting transactions that are fraudulent with great accuracy (98.46%) and high F1-score (97.63%). Where the ANN was the gold standard, the work outperformed it. Improvements in system quality have been achieved via feature engineering, SMOTE (a method for dealing with class imbalance), and optimization algorithm analysis. As for future research, there is a great opportunity to continue work on creating the system for real-time detection of frauds using stream processing methods. In addition, the application of explainable artificial intelligence techniques may be considered, which can help in ensuring regulatory compliance. The creation of adaptive learning strategies will also enable the system to remain in par with the changes in the way fraudsters operate. The other significant activity would be to engage multi-source data (behavioral data, graph analytics) in the process.